\documentclass{INTERSPEECH2023}

\interspeechcameraready 
\usepackage{cleveref}
\usepackage{ragged2e}

\title{Developing Speech Processing Pipelines for Police Accountability}
\name{Anjalie Field, Prateek Verma, Nay San, 
Jennifer L. Eberhardt, Dan Jurafsky}
\address{
  Stanford University, USA
  }
\email{\{anjalief, prateekv, nay.san, jleberhardt, jurafsky\}@stanford.edu}

\begin{document}

\maketitle
 
\begin{abstract}
Police body-worn cameras have the potential to improve accountability and transparency in policing.~Yet in practice, they result in millions of hours of footage that is never reviewed. We investigate the potential of large pre-trained speech models for facilitating reviews, focusing on ASR and officer speech detection in footage from traffic stops.~Our proposed pipeline includes training data alignment and filtering, fine-tuning with resource constraints, and combining officer speech detection with ASR for a fully automated approach.
We find that (1) fine-tuning strongly improves  ASR performance on officer speech (WER=12-13\%), (2) ASR on officer speech is much more accurate than on community member speech (WER=43.55-49.07\%), (3) domain-specific tasks like officer speech detection and diarization remain challenging. Our work offers practical applications for reviewing body camera footage and general guidance for adapting pre-trained speech models to noisy multi-speaker domains.

\end{abstract}

\section{Introduction}
Over the last decade, police departments across the United States have rapidly adopted body-worn cameras (BWCs) \cite{zansberg2021public}.
This rapid adoption has been spurred on by widespread protests demanding improved accountability and transparency following high-profile deaths of civilians involving officers' use of force \cite{lum2019research,elkins2023validating}.~In some ways, BWCs have resulted in improvements: the footage is valuable evidence in instances such as litigation of excessive force cases \cite{zamoff2019assessing,ccubukccu2021body}, and analysis of hand-transcribed footage can identify racial disparities in policing and failures to practice procedural justice \cite{voigt2017language,prabhakaran-etal-2018-detecting,rho23}.
However, in the absence of a lawsuit or high-profile incident, most footage is never reviewed. 
Further, reliance on manual transcriptions limits the scalability of existing automated analyses \cite{voigt2017language,camp2021,rho23}.

At the same time, large pre-trained speech models have achieved remarkable performance over standardized datasets \cite{baevski2020wav2vec,schneider2019wav2vec,hsu2021hubert,hsu2021robust,radford2022robust}.~Models like Whisper and Wav2Vec2 also have demonstrated potential in social good applications, e.g., in monitoring audio(visual) materials related to long-term elderly care \cite{Hacking2022} or child exploitation \cite{Vásquez-Correa2023}.~However, in applications involving multi-speaker conversations in noisy environments, models require application-specific adaptation and evaluation \cite{dutta2022challenges,abulimiti2020automatic,zuluaga2021bertraffic}. Little work has investigated the speech processing of police BWC footage specifically.

Here, we develop and evaluate automatic speech recognition (ASR) and police officer speech detection (diarization) for police BWC footage.~Automatic transcription of officer speech would allow extending existing text analyses of racial bias in hand-transcriptions to new data without requiring expensive transcription efforts \cite{voigt2017language,rho23}.
It would also allow departments to determine adherence to a procedure by using text classifiers \cite{prabhakaran-etal-2018-detecting} or keyword searches.
Although most reviews are likely to be internal, some departments publicly release BWC footage or are mandated to provide access upon request \cite{lin2015police,zansberg2021public}.
Thus, speech-processing technology could support independent audits.

Our primary data is footage from 1,040 vehicle stops conducted by one department in one month, where utterances spoken by officers and community members were previously hand-transcribed.
We use the data to construct training and test data sets for ASR and officer speech detection. We evaluate ASR models, with and without in-domain fine-tuning, over the entire test set, dividing by role (officer or community member), race, and gender, and we examine the performance of officer speech detection in combination with ASR.

Our findings provide insight into the best practices and limitations of developing technology in this domain. For example, our training data processing pipeline is robust enough that fine-tuning improves ASR performance by 3-11 points. We also show evidence that Whisper models learn to mimic transcribers' representations of transcription confidence by marking difficult segments as unintelligible. Differences by gender and race are not significant; however, ASR over officer speech (WER=12-13\% for officers unseen in training) is much more accurate than over community member speech (WER=43.55-49.07\%), which suggests that models have a high potential for addressing accountability with less risk of compromising community member privacy \cite{lin2015police}.
Finally, we identify diarization, specifically officer speech detection, as a continued challenge.

\section{Data}
\label{sec:data_processing}

Video recordings of the 1,040 vehicle stops and hand-transcriptions were provided to us under a data use agreement for the management of such high-risk data and under IRB supervision. The data is generally noisy. Prior transcripts were intended for language analysis, rather than the development of speech processing tools, so not all speech was transcribed and diarized.\footnote{The  transcribers were instructed to transcribe only speech by officers and community members, not  police dispatch;  they inconsistently included officer speech to dispatch (vs. to the community member).} 
Stops contain background noise like wind and traffic. They contain multiple speakers, and secondary officers, as well as drivers and passengers, can be situated far from the recording device. Dispatch speech from officers' radios can often be heard, sometimes directly overlapping with utterances from the primary interaction.
There is high variance in the clarity of speech and quality of footage across stops. 

\noindent {\bf Test and Validation Sets.} To create reliable test and validation sets, we hand-align existing transcribed utterances to \hyphenation{time-stamps} timestamps and correct observed transcription errors. To facilitate analysis by race, we chose the test data to consist of 50\%/50\% stops of white and black drivers. We also choose each test file to be a stop by a distinct officer and withhold any other stops made by the same officers (whether as primary or secondary officers) from the training and validation sets. Thus, we also selected officers who made a small number of stops to minimize unusable data. Hand-aligning data is extremely time consuming, so we restrict test set stops to contain $<60$ utterances. We similarly ensure there is no overlap in primary officers between the validation and training set, witholding data as needed, though we less strictly enforce the separation of secondary officers, who speak less frequently.
We conduct evaluations over these aligned utterances, discarding un-transcribed speech.

\begin{table}[]
    \centering
    \begin{tabular}{cccccc}
    \hline
              &  Robust   &  Whisper & Prop. of\\
    Alignment &  W2V2 WER &  Large WER &  Final data \\
    \hline
    \hline
    Unaligned    &  65.51 & 56.78 & 13.86 \\
    MFA          & 68.42 & 54.84 & 12.01 \\
    MFA Chunked  & 61.11 & 42.04 & 32.32 \\ 
    W2V2         & 60.25 & 43.27 & 12.72 \\ 
    W2V2 Chunked & 68.0 & 52.27 & 29.10 \\

    \end{tabular}
    \caption{WER over the full training set (78K utterances) under each alignment method and what percentage of training data were ultimately aligned with each method.}
    \label{tab:alignment_wer}
    \vspace{-5mm}
\end{table}

\noindent {\bf Training Set Alignment.} We build a training set by applying automated alignment tools and filtering poor-quality transcriptions.
We determine the start and end time for each transcribed utterance using the best of 5 alignment methods:
\begin{itemize}
    \item Unaligned: 1sec granularity timestamps hand-written by transcribers with heuristics to correct for obvious typos and extending the start and end by .25sec
    \item MFA: Montreal Forced Aligner \cite{mcauliffe2017montreal} with unaligned timestamps as starting points
    \item MFA chunked: Many utterances are too short for the aligner to process correctly. Thus, using the unaligned timestamps, we chunk consecutive utterances up to a total of 20sec. We run MFA to obtain word-level timestamps and then divide  chunks back into separate utterances, with start and end times determined by the word-level timestamps
    \item W2V2: Robust Wav2Vec2 \cite{hsu2021robust} for forced alignment \cite{kurzinger2020ctc}
    \item W2V2 chunked: Same as MFA chunked, but using Robust Wav2Vec2 for forced alignment instead of MFA.
\end{itemize}

For each utterance, we use off-the-shelf Whisper Large \cite{radford2022robust} and Robust Wav2Vec2 (W2V2) \cite{hsu2021robust} to transcribe the audio segment identified by each alignment method and compare the output with the hand-written transcript. We choose as the final alignment the one for which $min(WER_{Whisper}, WER_{W2V2})$ is lowest. \Cref{tab:alignment_wer} reports training WER for each alignment method and the percent of the final training data aligned using each method.

\noindent {\bf Training Set Filtering.} Even after alignment, the training data is noisy, containing, for example, transcription errors, overlapping speech, and unfixed alignment errors. We again use $min_{WER} = min(WER_{Whisper}, WER_{W2V2})$ over the best alignment to filter out training instances that are likely incorrect. We experiment with four filtering criteria, indicating filtered training data size in brackets:

\begin{enumerate}
    \item Remove instances $<0.5$sec and $>10$sec [54,600]
    \item \#1, and remove instances where $min_{WER} > 50\%$ [40,361]
    \item We define $WER[no subs.]$ as WER where we do not count substitutions as errors. This metric is designed to retain instances where there may be errors in the Whisper/Wav2Vec2 outputs (e.g., WER is high) but likely not alignment errors (e.g., WER is driven by substitutions rather than insertions or deletions). We then filter according to \#1, and keep only instances where ($min_{WER[no subs.]} < 10\%$ AND $min_{WER} < 50\%$). [26,121]
    \item \#1, and remove instances where $min_{WER} > 10\%$ [19,759]
\end{enumerate}

We compare each criteria by using the filtered training data to fine-tune Robust Wav2Vec2 and examining performance over the validation set. Criteria \#3 (WER=45.23)  and \#4 (WER=44.92) perform similarly and both outperform \#1 (WER=49.34) and \#2 (WER=48.75). We use \#3 when training subsequent models, favoring the criteria that keeps more training data. \Cref{tab:final_data_statistics} reports the final sizes for each data split.

\begin{table}[]
    \centering
    \begin{tabular}{ccccc}
    \hline
           & \# Stops & \# Utterances   &  Speech Time \\
     \hline
     \hline
     Train            &  795  & 78,082 & 61.85hr  \\
     Train (filtered) &  787  & 26,121 & 17.61hr \\
     Validation       &   8   & 373    & 21.24min   \\
     Test             &  20   & 634    & 32.41min   \\

    \end{tabular}
    \caption{Final data set sizes. Across the full data set, there are an average of 91.73 utterances and 3.2 speakers per stop.}
    \label{tab:final_data_statistics}
    \vspace{-5mm}
\end{table}

\section{ASR}
\label{sec:ASR}

We compare the performance of ASR models off-the-shelf and fine-tuned on the training data set constructed in \Cref{sec:data_processing}. We use two of the current best-performing and most popular architectures: Wav2Vec2 \cite{baevski2020wav2vec} and Whisper \cite{radford2022robust}.
For Wav2Vec2, we use the Robust model \cite{hsu2021robust}, which was pre-trained using a self-supervised objective on Libri-Light, 
CommonVoice, Switchboard, Fisher and fine-tuned for ASR on Switchboard. For Whisper, which was trained on 680,000 hours of multilingual and multitask data, we compare \textit{small},  \textit{medium}, and \textit{large} \cite{radford2022robust}. Thus, both models are intended to perform well in a variety of domains and over noisy data.
We describe the model training parameters in detail, including the use of decoder-only training for Whisper large due to compute constraints.

\subsection{Experimental Setup}

To fine-tune Wav2Vec2, we use model default parameters with learning rate=1e-5, weight decay=0.005, warmup steps=500, batch size=32.~We report performance with and without a 4-gram language model trained over the training data transcripts, implemented with KenLM and integrated with beam size=1500, lm weight=1.31 and word score=1.31.\footnote{lm weight and word score were tuned following the Bayesian optimization procedure in \cite{baevski2020wav2vec}.
We do no other hyperparameter tuning.}

For Whisper models without fine-tuning, we hard-code the task as transcription and the language as English. For fine-tuning, we use model default parameters with learning rate=1e-5, and warmup steps=500. Our experiments are conducted in a resource-constrained environment. Data protocols mandate that the footage be stored on a secure restricted-access server, which does not have sufficient GPU memory to fine-tune Whisper large, even with reduced batch size and precision. Thus, we experiment with freezing the encoder and just training the decoder as well as the inverse.
We use a batch size of 32 for Whisper small and 16 for medium and large.
Finally, as Whisper is prone to outputting repeated words and phrases, we remove any words from the model output if they occur $>10$ times.

As transcription norms  vary between corpora and the body-camera gold transcripts contain bracketed terms like \textit{[unintelligible]} and \textit{[laughter]}, we remove all terms in brackets and use the Whisper text normalizer on both the reference and model output before computing WER for all models (including Wav2Vec2 models).
For all models, we choose the checkpoint with the lower validation WER after 5 epochs and train using 1-2 A40 GPUs. Wav2Vec2 and Whisper small models trained in $<5$hrs; Whisper medium and large models trained in $<16$hrs.

\subsection{Results}
\subsubsection{Overall ASR}

\begin{table}[h!]
    \centering
    \begin{tabular}{ccccc}
    \hline
     Tuned Params & WER\textsubscript{S} & CER\textsubscript{S} & WER\textsubscript{L} & CER\textsubscript{L} \\
    \hline
    \hline
    None    & 34.75  &  24.86  & 33.07  &  28.47 \\
    Encoder & 34.30  &  23.53  & 22.82  & 17.58  \\
    Decoder & 28.12  &  20.07  & 22.26  & 16.86 \\
    Enc+Dec & 26.07  &  18.76  & -       & -   \\
    \end{tabular}
    \caption{Whisper Small (S) and Large (L) validation performance with encoder-only or decoder-only training}
    \label{tab:whisper_frozen_modules}
    \vspace{-5mm}
\end{table}

\Cref{tab:whisper_frozen_modules} reports validation results (reserving the test set for final configurations) of freezing either the encoder or decoder when fine-tuning Whisper large and small. For Whisper small, decoder-only tuning performs almost comparably to tuning the entire model (28.12 vs., 26.07), whereas tuning only the encoder performs less well (34.30). For Whisper large, freezing the encoder or decoder provides advantages over no fine-tuning, though decoder-only tuning converged faster (2 vs. 5 epochs). Subsequently, we use decoder-only training for the fine-tuned the Whisper large model.



\begin{table}[h!]
    \centering
    \begin{tabular}{lcc|lcc}
    \hline
    Wav2Vec2 & WER & CER        & Whisper     & CER     & WER   \\
    \hline
    \hline
    [None]    & 45.01 & 31.57   & Small        & 32.13  & 22.83 \\
    +LM       & 38.91 & 31.27   & Small+Tune   & 22.09  & 16.30 \\ 
    +Tune     & 42.20 & 26.05   & Med.         & 26.21  & 18.36  \\
    +Tune+LM  & 32.29 & 25.97   & Med.+Tune    & 23.47  & 17.78 \\
              &       &         & Large        & 29.60  & 22.35 \\
              &       &         & Large+Tune   & \textbf{18.33} & \textbf{13.61} \\
    \end{tabular}
    \caption{ASR Results over police test set.}
    \label{tab:asr_results}
    \vspace{-5mm}
\end{table}

\Cref{tab:asr_results} reports the overall WER and CER for each model. Whisper large with fine-tuning performs the best overall. Fine-tuning gives improves performance by 3-11pts across models.

\begin{table}[h!]
    \centering
    \begin{tabular}{p{0.95in}|p{0.73in}|p{0.9in}}
Reference & Whisper & Whisper(tuned) \\
\hline
\hline
\RaggedRight{Yeah. I know, I'm trying to--} & \RaggedRight{I'll turn it.} & \RaggedRight{Yeah. [unintelligible].}  \\
\hline
\RaggedRight{Yeah. [unintelligible] expired like la-- December.} & \RaggedRight{The fire started in December.} & \RaggedRight{Yeah. [unintelligible] expired like December.}   \\
\hline
\RaggedRight{[unintelligible].} & It's going to be a bad traffic. &  \RaggedRight{It's going to be a bad traffic.}  \\

\end{tabular}
\caption{Test outputs of fine-tuned Whisper large}
\label{tab:whisper_examples}
\end{table}

As Whisper is a  new model with yet-limited work on understanding model performance and fine-tuning effects, we highlight a few examples from the data in \Cref{tab:whisper_examples}.
In the original transcripts, transcribers mark segments they are unable to decipher as \textit{[unintelligible]}. While we removed all bracketed text when computing WER rate for fair comparison of off-the-shelf and fine-tuned models, examining Whisper outputs reveals that the fine-tuned model sometimes outputs \textit{[unintelligible]}. In some instances, the predicted \textit{[unintelligible]} exactly aligns with hand-transcription. However, we also find examples where Whisper hallucinates transcriptions for difficult content, whereas Wav2Vec2 more often does not produce output. After fine-tuning, Whisper hallucinations are particularly difficult to identify without referring back to the audio, as they often appear to be plausible statements in an interaction.

\subsubsection{Performance by officer/driver, gender, and race}

We examine model performance over sub-populations of the test data, specifically distinguishing between officers and community members, black and white people, and men and women. As there is high variance in model performance depending on the quality of footage from each stop, we use a mixed effects linear regression model. Each data point in the regression is a single utterance. The dependent variable is model WER for the utterance. Role (officer or community member), race, gender are fixed effects, and the specific stop is a random effect.

\begin{table}[]
    \centering
    \begin{tabular}{ccccc}
    \hline
              & W2V2 & W2V2      & Whisp. & Whisp.   \\
              &      & \footnotesize{tuned+LM} &          & \footnotesize{tuned} \\
    \hline
    \hline
    Role [Officer]   &  -.440*  & -.435*  &  -.791* & -.383* \\
    Race [Black]     &  -.028   & -.026  &  -.350  & -.034  \\ 
    Gender [F]       &   .106   &  .088  & .215    &  .091  \\ 

\hline
    CM Black [120]     &  83.67  &  66.53  & 66.53 & 43.55\\
    CM White [130]     &  88.45  &  74.02  & 75.05 & 49.07 \\ 
    Off. Black [175]   &  42.14  &  27.26  & 19.43 & 13.11 \\ 
    Off. White [166]   &  32.80  &  21.95  & 22.70 & 12.50 \\
    \end{tabular}
    \caption{
    ASR by role/race/gender for Robust W2V2 and Whisper Large (not including 3 Hispanic officers).
    Top: ASR Mixed Effects Regression. A negative (starred if significant) coefficient indicates lower WER (better performance).
    Bottom: WER for each subgroup. Brackets indicate number of test utterances}
    \label{tab:mixed_effects_coef}
    \vspace{-5mm}
\end{table}
\Cref{tab:mixed_effects_coef} reports the learned regression coefficients and WER by sub-population for the best performing Wav2Vec2 and Whisper models, off-the-shelf and fine-tuned. ASR performance for officers is significantly better than performance for community members by a wide margin. Even the best-performing models perform poorly at transcribing community member speech.
 Community members are situated further from the camera and typically speak very few short utterances.
Even hand-transcribers often mark their speech as unintelligible, and training a high-performing model on this type of data may be infeasible.
This result suggests that ASR could be an extremely useful tool for police accountability with small potential privacy-reducing impact on community members.

In contrast to prior work, we do not find significant differences by race or by gender \cite{Koenecke2020}. Subdividing the test data leads to small data set sizes, which could be skewed by a single outlying stop. This potential effect is greater when looking at race and gender than looking at role, since a low-quality video would decrease ASR performance for both the officer and the community member, whereas in examining race and gender, we are comparing across footage of different stops.
\Cref{tab:mixed_effects_coef} does show
WER is lower for white than black officers for most models.

\section{Officer Speech Detection}

In \Cref{sec:ASR}, we use hand-aligned evaluation data, but in practice, we do not know segmentation or speaker identities in new footage.
As our goal is police accountability, we develop two models to identify segments of speech by primary officers (e.g., officers wearing the camera) and evaluate them using the best-performing ASR model over the detected speech.

\subsection{Methodology}

\noindent \textbf{Training Data Processing}
We adapt the training set introduced in \Cref{sec:data_processing}. We remove any instances that do not contain active speech using an off-the-shelf acoustic scene understanding Mobile-Net \cite{howard2017mobilenets} architecture trained on AudioSet \cite{gemmeke2017audio} (AudioSet category 0 $<0.3$).
We divide remaining samples into 250ms chunks with a 100ms hop and represent each 250ms segment as a mel-spectogram with 64 mel-filters, computed with a hop of 10ms, and a window of 25ms. We create a balanced training corpus by randomly sampling 150K chunks each of officer/non-officer speech.
Since officers are closer to body-camera microphones (near-field) than community members (far-field), we use volume-based data augmentation.

As the raw training data contains non-officer speech that was not transcribed (e.g., dispatch speech), we also \textit{augment} the training set.
We divide training files into 250ms chunks with a 100ms hop, keep chunks with a speech score (from the Mobile-Net model) $\ge 0.5$, and merge consecutive chunks that occur within 1sec of each other. We add all new segments (ones that were not transcribed) to the training data as instances of not-officer-speech and then filter and sample the data as described above.
We use these data to train models to classify 250ms chunks as officer or not-officer speech (with cross-entropy loss).

\noindent \textbf{In-domain classifier}
We train a custom model from scratch, which contains 7 convolutional layers with 128 3x3 filters in every layer and Relu activation followed by max-pooling of 2. The output of the last layer is passed onto a linear head of 1024 neurons, followed by softmax activation, and the posterior probability is taken as officer score for that instance.

\noindent \textbf{Universal d-vectors} We extract d-vectors as features from an off-the-shelf model trained over the VoxCeleb dataset for speaker recognition \cite{wan2018generalized}  and train an officer speech classifier, with the same linear-head architecture as the in-domain model.

\noindent \textbf{Inference}
We predict voice activity detection (using the same
Mobile-Net model) and officer scores for 250ms chunks with
100ms hops. We consider a chunk to be officer speech if its voice activity score is $>t_{\text{VAD}}$ and its officer score is $>t_{\text{officer}}$, and we merge positive chunks if they occur within $t_{\text{smooth}}$ sec of each other.\footnote{\{$t_{\text{VAD}}$,$t_{\text{smooth}}$, $t_{\text{officer}}$\} are hyperparameters chosen via 20-iteration Bayesian optimization over the validation set with range [0,1] for $t_{\text{VAD}}$/$t_{\text{officer}}$ and [0.25,2] for $t_{\text{smooth}}$. They are \{0.93,1.76,0.16\} for d-vector, 
\{0.4,0.67,1.2\} for in-domain, and 
\{0.52,0.51,1.1\} for in-domain [aug.]}
For evaluation, we concatenate the ASR model output for all identified segments and compute WER against similarly concatenated hand-aligned officer segments.

\subsection{Results}

\Cref{tab:officer_detection_results} reports results for the best performing ASR model over the automatically detected officer speech segments. There is a substantial performance decrease between the hand-aligned segments and the detected segments.
The d-vector model performance particularly poorly, likely due to the high  difference in domain between VoxCeleb and police traffic stops.
Augmenting the training data does substantially improve performance (49.47 to 31.52 WER), though performance still may not be sufficient for applications.
In reviewing model outputs, we identify that models often misidentify other near-field speech as officer speech, and the presence of multiple officers complicates the task, as speech by secondary officers is sometimes scored closer to community member speech.
We also identified several annotation errors, such as segments attributed to the wrong person and inconsistencies in which speech was transcribed, suggesting these metrics may under-estimate performance.
These errors could be removed in hand-aligned test data, but their presence in training data is still likely to degrade model performance, and manually re-cleaning training data (as opposed to automatic augmentation) would involve a substantial undertaking that may not generalize to other settings.



\begin{table}[]
    \centering
    \begin{tabular}{cccccc}
    \hline
   Model & WER & CER & \%S & \%D & \%I  \\
    \hline
    \hline
     d-vector \cite{wan2018generalized}      & 61.85	& 53.14 & 15.50 & 	25.07 &	21.27 \\
     In-domain  & 49.47 & 39.83 & 15.15 & 21.10 & 13.22 \\
     In-domain [aug.] & 31.52 &	25.29 & 7.94 & 11.64 & 11.93 \\
     Hand-aligned  & 12.80  & 8.68  & 5.98  & 3.97  & 2.85 \\

    \end{tabular}
    \caption{ASR results over officer detected speech using tuned Whisper Large. S:substitutions, D:deletions, and I:insertions}\label{tab:officer_detection_results}
    \vspace{-10mm}
\end{table}


\section{Discussion}

We find pre-trained ASR models achieve low WER over police officer speech, particularly when fine-tuned on automatically cleaned training data.
Whisper specifically achieves low WER and even learns to mimic transcribers in marking segments as unintelligible, but can still fail more dramatically over difficult segments than Wav2Vec2.
While prior work has identified ASR as a limitation in noisy speech domains \cite{dutta2022challenges,abulimiti2020automatic}, we instead find that officer speech detection is a significant challenge in this setting. 
There are potential avenues for improvement, such as explicitly modeling dispatch and secondary officer speech or using text-based classifiers over ASR outputs \cite{zuluaga2021bertraffic}.
Further, although WER is worse over detected than hand-aligned officer speech, WER is an imperfect proxy metric for tasks actually of interest, such as determining officers' adherence to procedure.
As many errors are driven by misidentified or short utterances, performance may still be sufficient for tasks like dialog act classification \cite{prabhakaran-etal-2018-detecting}.
While we focus on policing, our work has the potential to inform adapting ASR models to other noisy multispeaker domains as well.

\noindent \textbf{Limitations and Ethical Considerations} Our data consists of traffic stops from one police department.
We cannot predict if results generalize to data from other departments, time periods or types of police-community interactions.
Also, although all work abides by IRB and data sharing protocols, it has high misuse potential. Models could used for purposes other than police accountability, such as community surveillance. Because models were trained on private data and pending mitigation of potential misuse, we are not releasing them at this time. 


\begin{center} \textbf{Acknowledgements} \end{center}
\vspace{-3mm}
\noindent This research was supported by the John D. and Catherine T. MacArthur Foundation (G-1512-150464 and G-1805-153038). We thank Stanford Data Science for fellowship funding as well as the city and the police department whose provision of camera footage enabled this research. This work was administered and supported by Stanford SPARQ, a center that builds research-driven partnerships to combat bias, reduce disparities, and drive culture change. We thank Martijn Bartelds, Dora Demszky, Rebecca Hetey, and Tolulope Ogunremi for helpful feedback.


\bibliographystyle{IEEEtran}
\bibliography{mybib}

\end{document}